%% file: template.tex
\documentclass{article}

\usepackage{arxiv}

\usepackage[utf8]{inputenc} 
\usepackage[T1]{fontenc}    
\usepackage{hyperref}       
\usepackage{url}            
\usepackage{booktabs}       
\usepackage{amsfonts}       
\usepackage{nicefrac}       
\usepackage{microtype}      
\usepackage{lipsum}
\usepackage{graphicx}
\usepackage{xcolor}
\usepackage{colortbl}
\usepackage{makecell}
\usepackage{array}
\usepackage{tabularx}
\usepackage{multirow} 
\graphicspath{ {./images/} }

\title{Automatic Feedback Generation for Short Answer Questions using Answer Diagnostic Graphs}

\author{
 Momoka Furuhashi \\
  Tohoku University, RIKEN (JAPAN)\\
  \texttt{furuhahashi.momoka.p4@dc.tohoku.ac.jp} \\
   \And
Hiroaki Funayama \\
  Tohoku University, RIKEN (JAPAN)\\
  \texttt{h.funa@dc.tohoku.ac.jp} \\
  \And
Yuya Iwase \\
  Tohoku University, RIKEN (JAPAN)\\
  \texttt{yuya.iwase.t8@dc.tohoku.ac.jp} \\
  \And
Yuichiroh Matsubayashi \\
Tohoku University, RIKEN (JAPAN)\\
  \texttt{y.m@tohoku.ac.jp} \\
  \And
Yoriko Isobe \\
RIKEN (JAPAN) \\
  \texttt{yoriko.isobe@riken.jp} \\
  \And
Toru Nagahama \\
Tohoku University (JAPAN) 
\\
  \texttt{toru.nagahama.a7@tohoku.ac.jp} \\
  \And
Saku Sugawara \\
National Institute of Informatics (JAPAN)\\
  \texttt{saku@nii.ac.jp} \\
\And
Kentaro Inui \\
MBZUAI (UAE), Tohoku University, RIKEN (JAPAN)\\
  \texttt{kentaro.inui@mbzuai.ac.ae} 
}

\begin{document}
\maketitle
\begin{abstract}
Short-reading comprehension questions are widely used in reading education to foster understanding of the structure of a prompt text. These questions typically require students to read a specific passage (prompt text) and then articulate their understanding of its contents in a few sentences. However, giving feedback to students on their responses to such problems can be burdensome for teachers. As a result, students generally only receive scores on their responses, making it difficult for them to identify and correct their own errors. Thus, it is a necessary to develop a system that automatically generates feedback statements, linking their responses to the scoring rubrics. Natural language processing (NLP) has evolved significantly in recent years. Automatic scoring feature remains a uniquely researched aspect in relation to short-reading comprehension questions, while feedback generation remains largely unexplored. To address this, we develop a system that can produce feedback for student responses. The Answer Diagnostic Graph (ADG) we proposed aligns the student’s responses to the logical structure of the reading text of these questions and automatically generates feedback. In our experiment, we assess the impact of our system using oracle feedback generated when the system is fully functional. The two experimental groups of students are asked to answer two prompts and their scores are compared: for these two prompts, one group receives the model answer and corresponding explanatory text (answer explanation condition) and the other receives our system’s oracle feedback in addition to those two (feedback condition), alternatively. We further investigated the students’ perceptions of the feedback and assess changes in their motivation. As a result, no significant differences were observed between the groups in terms of score improvements in re-answering. However, we found that feedback helped students understand the reasons for their mistakes and advance their comprehension of the key points of the text. We also found that feedback makes students enhance their motivation, but room remains for improvement in the generated feedback to promote understanding of the logical structure of the text.
\end{abstract}


\section{Introduction}
In reading education, short-reading comprehension questions are commonly utilized ~\cite{Snow2002ReadingFU}. 
These questions require students to read a given passage (prompt text) and respond to related prompts with several dozen words. 
The use of such method can improve student’s understanding of a text and develop their logical thinking ability ~\cite{Aloqaili2012TheRB}. 
However, this also presents practical issues in educational environments. 
First, this approach demands a considerable increase in teacher’s efforts in grading student responses and providing meaningful pedagogical feedback, especially when compared to multiple-choice questions ~\cite{bailey-meurers-2008-diagnosing}. 
Second, students’ responses often contain a range of errors, entailing the need for personalized feedback due to the varied scope they provide for areas for improvement. Nevertheless, students typically receive simple feedback that is focused only on their correctness with a numerical grade that does not explain the basis of their errors. 
This leaves students in the position of independently developing a way to improve their answers by comprehending the presented model answer and the corresponding explanatory text.

Recent advancement in natural language processing (NLP) have led to a growing interest in its educational applications.
One significant application of these technologies is the automated scoring of written responses for short reading comprehension exercises ~\cite{mizumoto-etal-2019-analytic, Sato2022, funayama2023reducing, Funayama2022}. 
While these existing studies have developed models to improve the accuracy of scoring, very little research has utilized NLP
techniques to generate feedback on these responses.
In this study, we aim to develop a system generating a personalized feedback comment on each student’s response to a short-reading comprehension question, and to assess its effectiveness through an empirical study with actual users.

To automatically generate personalized feedback for this question format, we propose a graph structure referred to as Answer Diagnosis Graphs (ADG) that integrates a directed graph representation of the logical structure of the target text with the templates of appropriate feedback comments by associating each template with a corresponding subgraph.
In this question format, the students are required to answer the question by referring to the relevant part of the text and summarizing the content.
The idea of our method is that if the referred part in the student’s response is not aligned with the appropriate part, the system identifies this misalignment on the ADG (i.e., graph representation of the text’s logical structure). 
The feedback is then generated depending on the logical relationship within this discrepancy, with the intension of leading students to notice this misalignment by themself. 
During our generation process of feedback, the system initially maps the student’s response to one of the ADG nodes, each of which corresponds to a sentence or a phrase in the target text. 
Subsequently, the ADG returns an appropriate feedback template attached to the identified node. 
Finally, the system produces feedback from the selected template using information derived from the student’s response text and the automated scoring model. 
To our best knowledge, it is the first system to generate personalized feedback for short-reading comprehension questions.

To evaluate the potential effectiveness of our proposed feedback generation method and identify any challenges associated with it, we conduct a practical experiment of our system with actual student users. 
In the experiment, Japanese high school students (n=39) are asked to answer two prompts, each of which require an answer in 70 to 80 words.
The students are divided into two groups, considering the difference in academic ability as little as possible, and for these two prompts, one group receives a model answer with the corresponding explanatory text, and the other group is provided with our system’s (oracle) feedback in addition to these two, alternatively. 
Then, both groups respond the prompt again, and changes in their scores are compared. 
An additional questionnaire is used to investigate the impression of our feedback and the changes in their motivation.

The results showed no significant difference in between two groups in terms of score improvement following re-answering. On the other hand, the result of our additional questionnaire showed that our feedback helped students understand the reasons for their mistakes and deepened their comprehension of the main points in the prompt text. 
It was also found that the feedback was significantly effective in increasing the student's motivation. 
However, room remains for improvement in generating feedback to promote understanding of the logical structure of the text.

\section{Ralated Work}
\label{sec:headings}
Research on generating automated feedback for students' response has been actively studied within the field of intelligent tutoring systems, but the main targets were STEM domains ~\cite{Paladines2020-vi}.
Meanwhile, in the field of language learning assistance, systems that generate feedback to improve students' writing abilities by identifying grammatical and spelling errors have widely studied ~\cite{Nur_Fitria2021-ad, Schmidt2022-lk}. 
There are also systems that provide feedback on essays where students express their opinions on provided topics ~\cite{xiao2024automation}. However, these systems assess the essay based on textual coherence and grammatical correctness, concentrating the feedback on a comprehensive evaluation of the entire text ~\cite{Paladines2020-vi, Wang2023-yg}. 
However, there is few research regarding personalized feedback for reading comprehension task, as addressed in this study. Our investigation of generating individualized feedback that facilitates a more logical interpretation of the learner's current comprehension marks a novel contribution to the field.

To evaluate the effectiveness of intelligent tutoring systems (ITSs), demonstrations in real educational settings are widely recognized as essential ~\cite{Goodman2001-xp}. 
ITSs have been tested for their effectiveness in various experimental contexts. 
These include providing random feedback that is not connected to students' responses, presenting subsequent problems based on learners' scores, and analyzing the learning outcomes through learning logs, as well as comparing the use of response-tailored feedback to generic feedback ~\cite{Rus2014-ev, Weerasinghe2010-zd, Graesser2004-jv}.
Our experimental setting mirrors these established paradigms. 
On the other hand, the primary means for assessing ITS is through measuring learning effectiveness. 
However, there exist limited studies investigating students’ motivation and engagement ~\cite{Wang2023-yg}.
In our study, we extend our scope beyond learning effectiveness to include the examination of students' motivation changes.
Most studies that have investigated engagement have used Likert-scale survey or learning logs ~\cite{Mohammadi_Zenouzagh2023-ym, Lin2023-yh, Veluvali2022-jk}, and we also adopt these methods in this study.

In recent years, rapid advances in Natural Language Processing (NLP) have led to a growing interest in research related to educational applications. One of the major areas of educational NLP is the
automated short answer scoring. 
Mizumoto et al. ~\cite{mizumoto-etal-2019-analytic} and Sato et al. ~\cite{Sato2022} output justification cues and scoring for responses. 
Funayama et al. ~\cite{Funayama2022} simulate a practical scenario where humans and an automatic scoring model grade responses in a cooperative way to further the deployment of automatic scoring models in the actual education field.
Han et al. ~\cite{han-etal-2024-llm} undertake automated scoring of essay papers and generated feedback using NLP technology.
However, there is few studies that have employed NLP technology to generate feedback on responses to short-reading comprehension questions. 
Hence, in this study, we aim to develop a system generating personalized feedback, and to assess its effectiveness through an empirical study with actual users.

\section{Dataset}
\begin{figure}[h]
    \centering
    \small
    \includegraphics[width=1\textwidth]{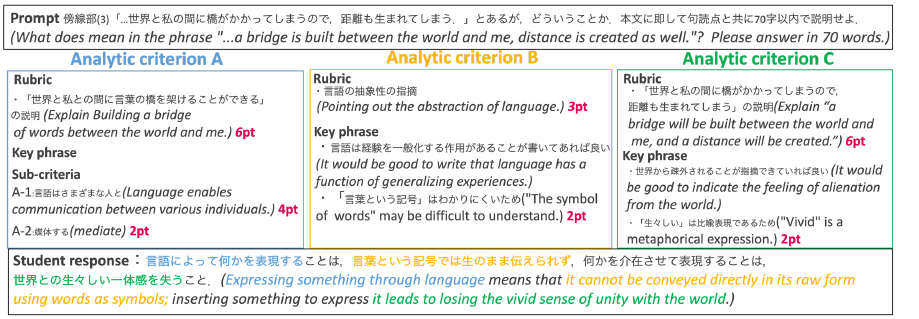} 
    \caption{An example of a prompt, analytic criteria, and a student response excerpted from the RIKEN SAS Dataset. Highlighted parts of the response indicate justification cues for each analytic score. We have omitted the prompt text in this example due to space limitations.}
    \label{fig:dataset}
\end{figure}

We use the RIKEN SAS Dataset introduced by Mizumoto et al. ~\cite{mizumoto-etal-2019-analytic} and expanded by Funayama et al. ~\cite{funayama2023reducing}. 
We show an example from this dataset in Fig. \ref{fig:dataset}.
The dataset contains pairs of responses and scores, with each response being graded by annotators.
Scoring rubrics are divided into several independent analytic criteria, and analytic scores are assigned based on these criteria. 
Substrings in the responses that provide rationale for the scoring are also annotated as justification cues.
Some of the analytic criteria are further subdivided into sub-criteria; however, the dataset does not include grades and justification cues for these sub-criteria.
Therefore, we additionally annotated the score and justification cues for these sub-criteria with the aim of generating elaborate feedback for them.
This dataset includes two types of text prompt: critical essays and narrative essays.
we use two prompts on critical essays out of the 13 problems included in the dataset.

To compare our proposed feedback design with a conventional reflective learning process, we utilize official answer explanations supplied by the company that created the prompts in our empirical study. 
The official explanations consist of a model answer and a detailed explanation of the reasoning process for each prompt.

\section{Feedback Generation}
\subsection{Feedback Generation task}
The reading comprehension prompts used in this study were designed to enable students to find information necessary for the correct answers within the given prompt text.
This allowed students to derive the appropriate responses from their understanding of the logical structure of the prompt text.
Consequently, we designed our feedback to highlight the gaps between students’ misunderstanding and the structures of the prompt text.
In addition, feedback that connects analytic criteria to students’ responses has been shown to be effective in reading education ~\cite{McGarrell2007}.

Considering these insights, our system generates feedback for every independent analytic criterion in the scoring rubric, which can help scaffold further understanding of the text structure.
Mizumoto et al. ~\cite{mizumoto-etal-2019-analytic} proposes an automatic scoring task that predicts analytic scores for each analytic criterion and identifies justifications cues associated with them.
We position our feedback generation system to work complementarily to the automatic scoring task.
Thus, we define feedback generation as a task that takes the students' response, an analytic score, and their justification cues outputted by the automated scoring model, and then provides feedback based on these elements.

\subsection{Our system}
\begin{figure}[h]
    \centering
    \small
    \includegraphics[width=1\textwidth]{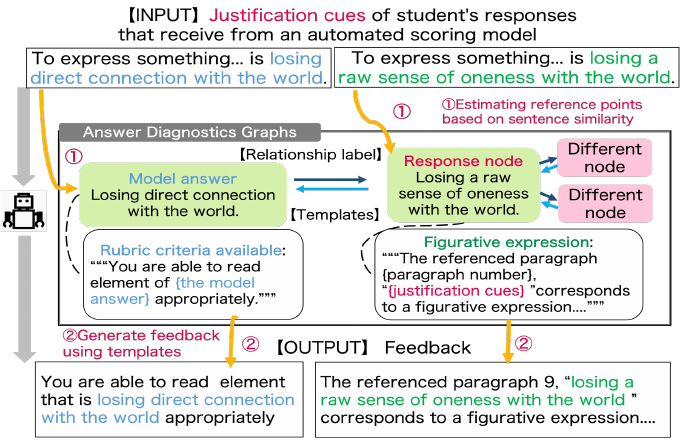} 
    \caption{Overview of Feedback Generation.}
    \label{fig:ADG_overview}
\end{figure}

The overview of our system is shown in Fig. \ref{fig:ADG_overview}. One critical component of our feedback generation system is the “Answer Diagnostic Graph (ADG)”. The ADG represents the logical relationships between sentences in the prompt text and the model answer as a graph structure.

In ADG, each node corresponds to a sentence in the prompt text or the model answer, and each edge represents the relationship between these texts. Each edge is linked to relationship labels that depict their logical relations. Each relationship labels are also linked to the pre-made feedback template.

During the feedback generation process, the system first estimates the response node to which the student’s response refers by calculating the similarity of each node to the justification cue outputted by the automatic scoring model.
Then, the system determines the templates to be used for feedback by examining the relationship label between the response node and the model answer node. 
In this section, we describe the design of such feedback in detail.

\subsection{Feedback Design}
\label{Feedback Design}
\subsubsection{Answer Diagnostic Graph}
As previously discussed, the prompts in this study require students to explain reasons or details for a specific sentence in the prompt text.
Consequently, responses may include excerpts, paraphrases, or summaries of prompt texts.
Incorrect responses typically consist of a transcription or paraphrase of the incorrect part of the prompt text. 
Therefore, we can identify error type of a student response by analyzing the logical relationship between the incorrect part referred by the response and the model answer.

\begin{figure}[h]
    \centering
    \small
    \includegraphics[width=1\textwidth]{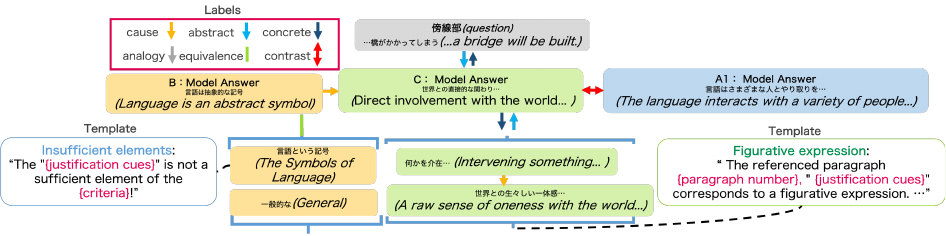} 
    \caption{An example of Answer an answer diagnosis graph. We created an ADG for each prompt text. A1, B, and C represent analytic criteria for the prompt.}
    \label{fig:ADG}
\end{figure}

To implement this, we developed the Answer Diagnostic Graph (ADG). 
An example of ADG is shown in Fig. \ref{fig:ADG}. 
This graph structure includes nodes representing the sentences of both the prompt text and the model answer, connected by edges that indicate their logical relationships.
Our design of the ADG aligns with previous research, which demonstrates that providing the logical structure of a text in graph format is effective for helping students grasp the structure of the text ~\cite{Li2020StudentsPO, Julia2017, Green2021CultivatingTS}.

To create the ADG, we first defined the relationship labels for the ADG based on rubrics, multiple textbooks, and the labels used in Rhetorical Structure Theory (RST) ~\cite{mann_1987}.
To create ADG, we manually split the prompt text into sentences.
We then further divided the sentences around the referred sentence in the prompt into chunks. We use these divided sentences or chunks as nodes.
We also add the justification cues of the model answer as nodes. Next, we manually assigned labels to each edge to indicate relationships between nodes.
Finally, we linked each edge to a template based on the relationship label of the edge.
These templates can include additional information such as paragraph numbers and answer hints for each node to provide detailed and personalized feedback for individual response.

\subsubsection{Template Construction}
As we constructed feedback templates based on an analysis of actual erroneous responses.
Specifically, we first categorized the types of erroneous responses in the development set into several common error patterns across the five prompts.
Then, we manually created oracle feedback for each error pattern. Finally, by aggregating those oracle feedback, we constructed ten generic feedback templates that can be applied across various prompts.
These templates can incorporate additional information, such as the paragraph number of the student’s response referred to, excerpts from the corresponding analytic criteria, and justification cues from the automatic scoring model to help students to understand their error or misunderstandings.

For example, in Fig. \ref{fig:ADG}, analytic criterion B require responses to imply "Language is an abstract symbol" to receive a full score.
However, if a response only refers to "Language is a symbol" and does not specify that the “symbol” is “abstract”, then the response is partially correct according to analytic criterion B.
In this case, the system use "Insufficient elements” template to generate feedback by incorporating the justification cue outputted by the automated scoring model and part of the analytic criterion B.
The generated feedback informs the student that certain elements are insufficient in their responses.

During the error analysis process, we also identified error types inherent to specific analytic criteria that do not fit into the generic template.
To address this issue, we created analytic templates based on such error types for some of the analytic criteria.
These analytic templates were created only for the two analytic criteria used in our experiments.

\subsection{Generation}
The Feedback for each analytic criterion was generated using a justification cue from the automatic scoring model.
For each response, we calculate the similarity between the justification cue and each node in the ADG using the Sentence-BERT ~\cite{reimers-2019-sentence-bert}.
We then identified the node with the maximum similarity as the response node for the response.
Feedback was generated using the template that was associated with the edge between the response node and the model answer node, inserting additional information described in the previous subsection.

\section{Empirical Study}
In the experiment, we evaluate the design of our feedback and potential of our feedback generation system.
We examine the effects of the feedback provided to students when an oracle is used for alignment between the student's response and the corresponding node in ADG described in Sec. \ref{Feedback Design}.

We assess our feedback design in terms of students’ understanding of the logical structure of a prompt text and their engagement, focusing on how it motivates students to correct their responses.
Furthermore, we collect qualitative feedback from participants to gather insights on how to improve our feedback design.

\subsection{Experimental Setting}
We validate our feedback by asking participants to respond the prompts again after reading the generated feedback.
41 Japanese high school students participate in our empirical study. They answer the prompts in 70-80 words.
The employed two prompts are designed for first-year and second-year high school students.
\begin{figure}[h]
    \centering
    \small
    \includegraphics[width=1\textwidth]{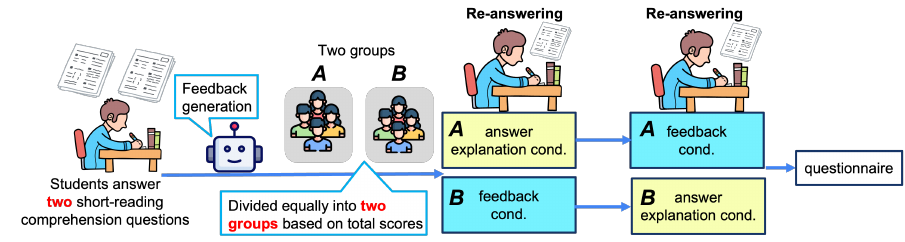} 
    \caption{Overall workflow of the experiment.}
    \label{fig:workflow_of_the_experiment.}
\end{figure}

The overall experimental workflow is illustrated in Fig. ~\ref{fig:workflow_of_the_experiment.}. The participants initially respond to each prompt and then engage in a re-answering phase.
In the initial phase, participants are required to answer two prompts, with 20 minutes allocated for each prompt.
Responses are collected via Google Forms and graded using an automated scoring system ~\cite{funayama2023reducing}.
Feedback for each analytic criterion is then generated by our system, based on the participants' responses and the scoring result.
As previously mentioned, our primary focus is on evaluating the feedback design and the potential of our system using ADG.
Therefore, we use the oracle matching results in the feedback generation with the oracle automatic scoring results. Finally, we integrate each analytical feedback and score to create comprehensive feedback for each student.

Building on previous research investigating the effectiveness of feedback ~\cite{Weerasinghe2010EvaluatingTE}, we conduct a comparative experiment with two distinct conditions in the re-answering phase: the provision of the official explanations (the answer explanation condition) and the delivery of supplemental feedback through our system (the feedback condition).
The participants are divided into two groups based on their total score from the initial phase, ensuring an unbiased distribution of academic ability across both groups. The first group is introduced to the answer explanation condition for the first prompt and to the feedback condition for the second prompt (and vice versa for the second group).
This setup allows both groups to re-answer the prompts under each individual condition. 25 minutes are allocated for re- answering each prompt. Following the re-answering phase, all participants are required to fill out multiple-choice questionnaires, detailed further in in the next section.

Following the re-answering phase, participants are instructed to complete multiple-choice questionnaires and a free-text comment form.
The multiple-choice questionnaires consist of three types: Type 1 asks absolute scores on usefulness of our feedback and the official explanation; Type 2 asks to compare the two conditions to assess which one is more effective for deeper understanding; and Type 3 gathers participants' impressions on whether our feedback met its intended objectives.

\subsection{Results}
Table \ref{type1_result} presents the questions and the results of the absolute-scoring questionnaire (Type 1).
In this questionnaire, participants were asked to respond using a five-point Likert scale for each question.
We show the average and standard deviation based on 35 responses for each question, along with the t- test results. The results indicate that the feedback condition elicited significantly more favourable responses than the answer explanation condition for six of the eight questions. 
Conversely, the remaining two questions did not yield statistically significant results.
\input{table/type1_result}

Table \ref{type2_result} displays the questions and the results of the comparison questionnaire (Type 2), where participants were asked to determine which condition was more suitable for deeper understanding.
Unexpectedly, the results revealed contradictory findings for two questions.
While most participants found the provided feedback useful for grasping the main points when re-answering the questions, the majority of participants found it easier to understand the logical structure of the prompt texts with only the official answer explanations.
\input{table/type2_result}

Table \ref{type3_result} shows the questions and the results of the chi-square test for the Type 3 questionnaire.
In this questionnaire, participants were asked to rate their impression of the provided feedback across four aspects: “Individualization”, “Relevance”, “Degree of Demand”, and “Learning Progression” using a six- points scale.
For analytical purposes, the scale responses were further divided into three groups: negative, neutral, and positive. 
The chi-square test revealed statistically significant differences between negative and positive responses and between neutral and positive responses for “Individualization”, “Degree of Demand” and “Learning Progression”.
Additionally, there was a statistically significant difference between the negative and neutral responses for “Relevance”; however, no significant difference was found between the neutral and positive responses.
\input{table/type3_result}

Participants were also asked to provide free-text comments on the feedback, resulting in a total of 57 comments.
These comments were categorized into positive and negative groups and further subcategorized.
Positive comments included “Improvements in responses”, “Points of the prompt text”, and “High understandability”. Negative comments included “Difficulty in understandability”, “Ambiguous instruction for revision”, and “Others”.
Table \ref{type4_result} shows the number of comments in each category. We received 20 comments on “Improvements in responses”, indicating that our feedback was useful material for identifying areas needing improvement in their responses. However, ten comments on “Ambiguous instruction for revision” suggest that some participants found the feedback too vague, highlighting limitations in our feedback design based on templates.
\input{table/type4_result}

After the explement, we re-graded the responses from the re-answering phase.
We conduct a t-test to analyse the scores between the initial and re-answered responses for the two prompts.
However, the results, shown in Table \ref{type5_result}, indicate no significant difference between the two conditions.
\input{table/type5_result}

\section{Conclusion}
In reading education, some research has demonstrated that personalized feedback significantly promotes students’ comprehension and motivation.
However, delivering such feedback imposes a substantial burden on educators. This study addressed this challenge by developing a feedback generation system that employes a novel graph structure called Answer Diagnosis Graph (ADG).
ADG represents the logical relationships between sentences in the prompt text used in reading comprehension questions.
It identifies gaps in understanding between students’ responses and the model answer, then provides suitable feedback templates for our system.

In the empirical study, we evaluated the potential educational effectiveness of our feedback generation system using oracle matching results to map student responses to the ADG.
The experimental results revealed that our feedback design encouraged students to objectively revise their own responses, identify the main structure of the prompt text, and positively influenced their emotional aspects, such as motivation and satisfaction. The indirect hints provided in our feedback facilitated students’ improvement of their responses rather than simply copying the model answer. During this revision process, the students identified the missing or insufficient elements in their responses or thinking process.
Consequently, as indicated by the Type 1 and 3 questionnaires, many of the participants were able to satisfactorily revise their responses and expressed a desire to use this feedback again in future learning.

Our investigation also revealed challenges with our feedback design.
The results from Type 1 and Type 2 questionnaires indicated that many participants found the current feedback too vague and struggled to understand how to revise their responses.
These findings highlight the need for improvements in the clarity and understandability of our feedback design.

\section{Limitations and Future work}
We acknowledge three limitations in this study.
First, we used oracle response node estimations and scoring results to generate feedback with ADG.
In practice, our system predicts response nodes from justification cues produced by an automatic scoring model. 
However, the accuracy of these predictions is insufficient for our empirical study and may introduce noise when evaluating the impact of our feedback in actual educational environment.
Therefore, we focused on evaluating our feedback design using oracle response node estimations and scoring results.
We plan to enhance the technological aspect of our system to achieve sufficient performance for the future empirical studies.

Second, the evaluation of emotional aspects based on the questionnaires tend to be biased.
Additionally, an overall positive bias may have emerged because the novelty of the feedback may have triggered curiosity among participants.
Our empirical study also lacks a quantitative analysis of the effectiveness of our feedback, such as its impact on learning outcomes.
Furthermore, our experimental condition might have distorted the results as participants could copy the model answer and re-answer the prompts due to the inclusion of the model answer in the official answer explanations.

To evaluate feedback design in more rigorously and from a pedagogical perspective, it is necessary to refine the experimental setup by incorporating insights from previous research.
For example, Jackson et.al measured the effectiveness of feedback by having participants respond to the same prompts after a period time ~\cite{Jackson2004TheIO}.
Li and Keller reported that many studies that have developed affective questionnaires based on the ARCS model to investigate affective aspects of feedback ~\cite{li2018use}.
In the future, we intend to include evaluation indicators such as the ARCS model in our questionnaires and conduct posteriori experiments.
We believe that this comprehensive investigation of our feedback design will provide valuable insights to improve our feedback system and design.

\section*{Acknowledgments}
We would like to thank Yamagata Prefectural Education Center and Yamagata Prefectural Sakata Koryo High School for their participation in our experiment.
We also wish to express our gratitude to Takamiya Gakuen Yoyogi Seminar for providing valuable rubrics, prompts and responses.
This work was supported by JSPS KAKENHI Grant Number JP22H00524, JST SPRING, Grant Number JPMJSP2114.

\bibliographystyle{unsrt}
\bibliography{references}

\end{document}

%% file: table/type1_result.tex
\begin{table}[h]
    \centering
    \caption{Type 1 Questions Survey Results.}
    \label{type1_result}
    \begin{tabular}{|>{\raggedright}p{8.5cm}|c|c| c|c|c|}
        \toprule
\rowcolor{gray!20}
        \textit{Question items} & \multicolumn{2}{c}{\textbf{Answer exp. cond.}} & \multicolumn{2}{c}{\textbf{Feedback cond.}} & \textbf{T-test} \\
\rowcolor{gray!20}
        & \textbf{Average} & \textbf{S.D.} & \textbf{Average} & \textbf{S.D.} & \textbf{results} \\
        \midrule
        I was more motivated to re-answer. & 4.2 & 1.4 & 5.2 & 0.7 & ** \\
        I could recognize good points in my responses. & 3.5 & 1.2 & 4.7 & 1.2 & ** \\
        I could recognize errors in my responses. & 4.8 & 1.2 & 5.5 & 0.7 & ** \\
        I could understand what to focus on when re-answering my responses. & 4.5 & 1.3 & 5.4 & 0.8 & ** \\
        I could find helpful explanations when re-answering my responses. & 5.0 & 1.3 & 5.6 & 0.8 & * \\
        I could easily understand the logical relationship of the prompt text. & 4.4 & 1.0 & 4.6 & 0.9 & ns \\
        I could gain confidence when re-answering. & 4.1 & 1.3 & 4.5 & 1.0 & ns \\
        I could satisfactorily re-answer my responses. & 4.3 & 1.2 & 4.8 & 0.9 & ** \\
        \bottomrule
    \end{tabular}
    \vspace{1mm}
    
    {\footnotesize **: (p$<$0.01), *: (p$<$0.05), ns: no significant difference, S.D.: standard deviation}
\end{table}

%% file: table/type2_result.tex
\begin{table}[h]
    \centering
    \caption{Type 2 Questions Survey Results.}
    \label{type2_result}
    \renewcommand{\arraystretch}{1.2}
    \setlength{\tabcolsep}{5pt} 
    \begin{tabular}{|
        >{\raggedright\arraybackslash}p{6cm}|  
        >{\centering\arraybackslash}p{2cm}|    
        >{\centering\arraybackslash}p{2cm} |   
        >{\centering\arraybackslash}p{2.5cm} |
        >{\centering\arraybackslash}p{2cm}|    
    }
        \hline
\rowcolor{gray!20}
        \textit{Question item} & \textit{Answer exp. cond. (I)} & \textit{Feedback cond. (II)} & \textit{Neither remains the same cond. (III)} & \textit{Chi-square result} \\
        \midrule
        I found it easier to understand the logical relationship between sentences when I was re-answering. & 22 & 8 & 5 & p$<$0.01 (I) \\
        \hline
        I found it easier to grasp the point when solving and re-answering. & 2 & 32 & 1 & p$<$0.01 (II) \\
        \bottomrule
    \end{tabular}
\end{table}

%% file: table/type3_result.tex
\begin{table}[h]
    \centering
    \renewcommand{\arraystretch}{1.2}
    \setlength{\tabcolsep}{3pt} 
    \caption{\textit{Type 3 Questions Survey Results.} $\times$, $\triangle$, $\circ$ denote negative, neutral, positive, respectively.}
    \label{type3_result}
    \begin{tabular}{|p{5cm}|>{\centering\arraybackslash}p{1.2cm}|>{\centering\arraybackslash}p{1.2cm}|
    >{\centering\arraybackslash}p{1.2cm}|>{\centering\arraybackslash}p{1.2cm}|
    >{\centering\arraybackslash}p{1.2cm}|>{\centering\arraybackslash}p{1.2cm}|
    >{\centering\arraybackslash}p{2.5cm}|}
        \hline
        \rowcolor{gray!20}
        \textit{Question} & \multicolumn{2}{c|}{Negative} & \multicolumn{2}{c|}{Neutral} & \multicolumn{2}{c|}{Positive} & \textit{Results} \\
        \rowcolor{gray!20}
        & \textit{Strongly disagree} & \textit{Mostly disagree} & \textit{Not so much agree} & \textit{Somewhat agree} & \textit{Agree} & \textit{Strongly agree} & \\
        \hline
        \rowcolor{gray!30} 
        \multicolumn{8}{|c|}{\textbf{Individualization}} \\
        \hline
        The feedback was generated in accordance with my response. & 0 & 2 & 1 & 3 & 13 & 16 & 
        \makecell{$\times=\triangle$: ns \\ $\times<\circ$:** \\ $\triangle<\circ$:**} \\
        \hline
        \rowcolor{gray!30}
        \multicolumn{8}{|c|}{\textbf{Relevance}} \\
        \hline
        The feedback was aligned with my level of understanding. & 0 & 1 & 5 & 7 & 12 & 10 & 
        \makecell{$\times<\triangle$:** \\ $\times<\circ$:** \\ $\triangle=\circ$: ns} \\
        \hline
        \rowcolor{gray!30}
        \multicolumn{8}{|c|}{\textbf{Degree of Demand}} \\
        \hline
        I wanted to use the feedback with answer explanations when reviewing. & 0 & 0 & 0 & 4 & 12 & 19 & 
        \makecell{$\times=\triangle$: ns \\ $\times<\circ$:** \\ $\triangle<\circ$:**} \\
        \hline
        \rowcolor{gray!30} 
        \multicolumn{8}{|c|}{\textbf{Learning Progression}} \\
        \hline
        Continuing to revise based on the system feedback could lead to a better understanding of how to interpret critical essays. & 0 & 0 & 2 & 5 & 14 & 14 & 
        \makecell{$\times<\triangle$:** \\ $\times<\circ$:** \\ $\triangle<\circ$:**} \\
        \hline
    \end{tabular}
    \vspace{3pt}
    \flushleft{**: (p$<$0.01), ns: no significant difference}
\end{table}

%% file: table/type4_result.tex
\begin{table}[h]
    \centering
    \caption{Categorization of Free-text Responses. All response examples are translated from Japanese.}
    \label{type4_result}
    \renewcommand{\arraystretch}{1.2}
    \setlength{\tabcolsep}{3pt} 
    \begin{tabular}{|l|p{4cm}|c|p{7cm}|}
        \hline
        \rowcolor{gray!20}
        \textbf{Category} & \textbf{Subcategory} & \textbf{\#Responses} & \textbf{Response example} \\
        \hline
        \multirow{4}{*}{Positive} & Improvements in responses & 20 & The feedback highlighted the insufficient elements in my responses, which were not in the official explanation. \\
        \cline{2-4}
        & Points of the prompt text & 6 & I can easily grasp the main points of the prompt text. \\
        \cline{2-4}
        & High understandability & 5 & The feedback told me the information I needed clearly and concisely. \\
        \cline{2-4}
        & Other & 4 & The feedback provided an explanation that aligned with my response. \\
        \hline
        \multirow{3}{*}{Negative} & Difficulty in understandability & 5 & I felt the feedback was slightly confusing. \\
        \cline{2-4}
        & Ambiguous instruction for revision & 10 & I needed more detailed information about which part of my response needs revision. \\
        \cline{2-4}
        & Other & 7 & I wanted to see an example of a revision of my responses. \\
        \hline
    \end{tabular}
\end{table}

%% file: table/type5_result.tex
\begin{table}[h]
    \centering
    \caption{\textit{Comparison of Score Improvement after Re-answering.}}
    \label{type5_result}
    \renewcommand{\arraystretch}{1.2}
    \setlength{\tabcolsep}{3pt} 
    \begin{tabular}{|c|c|c|c|c|c|}
        \hline
        \rowcolor{gray!20}
        \textbf{\textit{Prompt}} & \textbf{\textit{Condition}} & \textbf{\textit{\#Responses}} & \textbf{\textit{Mean}} & \textbf{\textit{S.D.}} & \textbf{\textit{T-test result}} \\
        \hline
        \multirow{2}{*}{1} & Feedback & 20 & 5.9 & 4.28 & \multirow{2}{*}{ns} \\
        \cline{2-5}
        & Answer explanation & 19 & 5.79 & 4.47 & \\
        \hline
        \multirow{2}{*}{2} & Feedback & 19 & 2.53 & 4.22 & \multirow{2}{*}{ns} \\
        \cline{2-5}
        & Answer explanation & 20 & 4.05 & 3.04 & \\
        \hline
    \end{tabular}
    
    \vspace{0.5em}
    \raggedright ns: No significant difference, S.D.: standard deviation. \\
    Both of questions were worth 15 points.
\end{table}

%% file: template.bbl
\begin{thebibliography}{10}

\bibitem{Snow2002ReadingFU}
Catherine~E. Snow.
\newblock Reading for understanding: Toward an r\&d program in reading comprehension.
\newblock 2002.

\bibitem{Aloqaili2012TheRB}
Abdulmohsen~S. Aloqaili.
\newblock The relationship between reading comprehension and critical thinking: A theoretical study.
\newblock {\em Journal of King Saud University - Languages and Translation}, 24:35--41, 2012.

\bibitem{bailey-meurers-2008-diagnosing}
Stacey Bailey and Detmar Meurers.
\newblock Diagnosing meaning errors in short answers to reading comprehension questions.
\newblock In Joel Tetreault, Jill Burstein, and Rachele De~Felice, editors, {\em Proceedings of the Third Workshop on Innovative Use of {NLP} for Building Educational Applications}, pages 107--115, Columbus, Ohio, June 2008. Association for Computational Linguistics.

\bibitem{mizumoto-etal-2019-analytic}
Tomoya Mizumoto, Hiroki Ouchi, Yoriko Isobe, Paul Reisert, Ryo Nagata, Satoshi Sekine, and Kentaro Inui.
\newblock Analytic score prediction and justification identification in automated short answer scoring.
\newblock In {\em Proceedings of the Fourteenth Workshop on Innovative Use of NLP for Building Educational Applications}, August 2019.

\bibitem{Sato2022}
Tasuku Sato, Hiroaki Funayama, Kazuaki Hanawa, and Kentaro Inui.
\newblock Plausibility and faithfulness of feature attribution-based explanations in automated short answer scoring.
\newblock In Maria~Mercedes Rodrigo, Noburu Matsuda, Alexandra~I. Cristea, and Vania Dimitrova, editors, {\em Artificial Intelligence in Education}, pages 231--242, Cham, 2022. Springer International Publishing.

\bibitem{funayama2023reducing}
Hiroaki Funayama, Yuya Asazuma, Yuichiroh Matsubayashi, Tomoya Mizumoto, and Kentaro Inui.
\newblock Reducing the cost: Cross-prompt pre-finetuning for short answer scoring.
\newblock In {\em International conference on artificial intelligence in education}, pages 78--89. Springer, 2023.

\bibitem{Funayama2022}
Hiroaki Funayama, Tasuku Sato, Yuichiroh Matsubayashi, Tomoya Mizumoto, Jun Suzuki, and Kentaro Inui.
\newblock Balancing cost and quality: An exploration of human-in-the-loop frameworks for automated short answer scoring.
\newblock In {\em Artificial Intelligence in Education}, pages 465--476, Cham, 2022. Springer International Publishing.

\bibitem{Paladines2020-vi}
José Paladines and Jaime Ramirez.
\newblock A systematic literature review of intelligent tutoring systems with dialogue in natural language.
\newblock {\em IEEE Access}, 8:164246--164267, 2020.

\bibitem{Nur_Fitria2021-ad}
Tira Nur~Fitria.
\newblock ``grammarly'' as {AI}-powered english writing assistant: Students' alternative for english writing.
\newblock {\em Metathesis Journal of English Language Literature and Teaching}, 5:65--78, 2021.

\bibitem{Schmidt2022-lk}
Torben Schmidt and T~Strassner.
\newblock Artificial intelligence in foreign language learning and teaching.
\newblock {\em Anglistik}, 33:165--184, 2022.

\bibitem{xiao2024automation}
Changrong Xiao, Wenxing Ma, Sean~Xin Xu, Kunpeng Zhang, Yufang Wang, and Qi~Fu.
\newblock From automation to augmentation: Large language models elevating essay scoring landscape.
\newblock {\em arXiv preprint arXiv:2401.06431}, 2024.

\bibitem{Wang2023-yg}
Huanhuan Wang, Ahmed Tlili, Ronghuai Huang, Zhenyu Cai, Min Li, Zui Cheng, Dong Yang, Mengti Li, Xixian Zhu, and Cheng Fei.
\newblock Examining the applications of intelligent tutoring systems in real educational contexts: A systematic literature review from the social experiment perspective.
\newblock {\em Educ Inf Technol (Dordr)}, pages 1--36, January 2023.

\bibitem{Goodman2001-xp}
P~Goodman.
\newblock Technology enhanced learning: Opportunities for change.
\newblock 2001.

\bibitem{Rus2014-ev}
Vasile Rus, Dan Stefanescu, William Baggett, Nobal Niraula, Don Franceschetti, and Arthur~C Graesser.
\newblock Macro-adaptation in conversational intelligent tutoring matters.
\newblock In {\em Intelligent Tutoring Systems}, pages 242--247. Springer International Publishing, 2014.

\bibitem{Weerasinghe2010-zd}
Amali Weerasinghe, Antonija Mitrovic, Martin Van~Zijl, and Brent Martin.
\newblock Evaluating the effectiveness of adaptive tutorial dialogues in database design.
\newblock 2010.

\bibitem{Graesser2004-jv}
Arthur~C Graesser, Shulan Lu, George~Tanner Jackson, Heather~Hite Mitchell, Mathew Ventura, Andrew Olney, and Max~M Louwerse.
\newblock {AutoTutor}: a tutor with dialogue in natural language.
\newblock {\em Behav. Res. Methods Instrum. Comput.}, 36(2):180--192, May 2004.

\bibitem{Mohammadi_Zenouzagh2023-ym}
Zohre Mohammadi~Zenouzagh, Wilfried Admiraal, and Nadira Saab.
\newblock Learner autonomy, learner engagement and learner satisfaction in text-based and multimodal computer mediated writing environments.
\newblock {\em Educ Inf Technol (Dordr)}, pages 1--41, April 2023.

\bibitem{Lin2023-yh}
Ying-Lien Lin and Wei-Tsong Wang.
\newblock Enhancing students' online collaborative {PBL} learning performance in the context of coauthoring-based technologies: A case of wiki technologies.
\newblock {\em Educ Inf Technol (Dordr)}, pages 1--26, June 2023.

\bibitem{Veluvali2022-jk}
Parimala Veluvali and Jayesh Surisetti.
\newblock Learning management system for greater learner engagement in higher {Education—A} review.
\newblock {\em Higher Education for the Future}, 9(1):107--121, January 2022.

\bibitem{han-etal-2024-llm}
Jieun Han, Haneul Yoo, Junho Myung, Minsun Kim, Hyunseung Lim, Yoonsu Kim, Tak~Yeon Lee, Hwajung Hong, Juho Kim, So-Yeon Ahn, and Alice Oh.
\newblock {LLM}-as-a-tutor in {EFL} writing education: Focusing on evaluation of student-{LLM} interaction.
\newblock In Sachin Kumar, Vidhisha Balachandran, Chan~Young Park, Weijia Shi, Shirley~Anugrah Hayati, Yulia Tsvetkov, Noah Smith, Hannaneh Hajishirzi, Dongyeop Kang, and David Jurgens, editors, {\em Proceedings of the 1st Workshop on Customizable NLP: Progress and Challenges in Customizing NLP for a Domain, Application, Group, or Individual (CustomNLP4U)}, pages 284--293, Miami, Florida, USA, November 2024. Association for Computational Linguistics.

\bibitem{McGarrell2007}
Hedy McGarrell and Jeff Verbeem.
\newblock Motivating revision of drafts through formative feedback.
\newblock {\em ELT Journal}, 61(3):228--236, 07 2007.

\bibitem{Li2020StudentsPO}
Liang-Yi Li and Chin-Chung Tsai.
\newblock Students’ patterns of accessing time in a text structure learning system: relationship to individual characteristics and learning performance.
\newblock {\em Educational Technology Research and Development}, 68:2569 -- 2594, 2020.

\bibitem{Julia2017}
Julia~V. Roehling, Michael Hebert, J.~Ron Nelson, and Janet~J. Bohaty.
\newblock Text structure strategies for improving expository reading comprehension.
\newblock {\em The Reading Teacher}, 71(1):71--82, 2017.

\bibitem{Green2021CultivatingTS}
Jennifer~M Green and Jennifer Holman.
\newblock Cultivating the strategy of summarizing sequential expository text: Scaffolds and supports for the intermediate grades.
\newblock {\em Literacy Practice and Research}, 2021.

\bibitem{mann_1987}
William~C. Mann and Sandra~A. Thompson.
\newblock Rhetorical structure theory: A theory of text organization.
\newblock Technical Report ISI/RS-87-190, Information Sciences Institute, June 1987 1987.

\bibitem{reimers-2019-sentence-bert}
Nils Reimers and Iryna Gurevych.
\newblock Sentence-bert: Sentence embeddings using siamese bert-networks.
\newblock In {\em Proceedings of the 2019 Conference on Empirical Methods in Natural Language Processing}. Association for Computational Linguistics, 11 2019.

\bibitem{Weerasinghe2010EvaluatingTE}
Amali Weerasinghe, Antonija Mitrovic, Martin van Zijl, and Brent Martin.
\newblock Evaluating the effectiveness of adaptive tutorial dialogues in eer-tutor.
\newblock 2010.

\bibitem{Jackson2004TheIO}
G.~Tanner Jackson, Matthew Ventura, Preeti Chewle, and Arthur~C. Graesser.
\newblock The impact of why/autotutor on learning and retention of conceptual physics.
\newblock In {\em International Conference on Intelligent Tutoring Systems}, 2004.

\bibitem{li2018use}
Kun Li and John~M Keller.
\newblock Use of the arcs model in education: A literature review.
\newblock {\em Computers \& Education}, 122:54--62, 2018.

\end{thebibliography}
